\pdfoutput=1
\documentclass[pdflatex,sn-mathphys-num]{sn-jnl}
\usepackage{graphicx}
\usepackage{float}
\usepackage{amsmath,amssymb}
\usepackage{booktabs}
\makeatletter\newcommand{\partialrule}[1]{\@cline#1\@nil}\makeatother %
\usepackage{lineno}
\def\doiurl#1{} %
\hypersetup{bookmarksdepth=subsection} %
\usepackage{etoolbox}
\makeatletter
\def\addressfont{\reset@font\fontsize{9bp}{11bp}\selectfont\titraggedcenter} %
\patchcmd{\@maketitle}{\vskip20pt\artauthors}{\vskip14pt\artauthors}{}{} %
\patchcmd{\@maketitle}{\removelastskip\vskip24pt}{\removelastskip\vskip10pt}{}{} %
\patchcmd{\@maketitle}{\removelastskip\vskip24pt}{\removelastskip\vskip14pt}{}{} %
\renewcommand\abstracthead{\@startsection{section}{1}{\z@}{-14pt \@plus0ex \@minus0ex}{3pt}{\abstractheadfont}} %
\makeatother

\begin{document}

\title[ConRef: Consensus Reference]{ConRef: Auditing Reference Reliability in 2D/3D Registration Benchmarks}

\author*[1]{\fnm{Jianfeng} \sur{Yue}}\email{jianfeng.yue@tum.de}
\author*[1,2]{\fnm{Feng} \sur{Li}}\email{feng.li@tum.de}
\author[1,2]{\fnm{Tianyu} \sur{Song}}
\author[3,4]{\fnm{Julius} \sur{Watrinet}}
\author[1]{\fnm{Ulrich} \sur{Eck}}
\author[1,2]{\fnm{Nassir} \sur{Navab}}
\affil[1]{\orgdiv{Chair for Computer-Aided Medical Procedures and Augmented Reality}, \orgname{Technical University of Munich}, \orgaddress{\city{Munich}, \country{Germany}}}
\affil[2]{\orgname{Munich Center for Machine Learning}, \orgaddress{\city{Munich}, \country{Germany}}}
\affil[3]{\orgname{BG Unfallklinik Murnau}, \orgaddress{\city{Murnau am Staffelsee}, \country{Germany}}}
\affil[4]{\orgdiv{Department of Orthopaedic Sports Medicine}, \orgname{Technical University of Munich}, \orgaddress{\city{Munich}, \country{Germany}}}

\abstract{
\textbf{Purpose:} 
Accurate 2D/3D registration is crucial for image-guided orthopedic interventions, enabling the alignment of preoperative 3D anatomy with intraoperative X-ray observations for surgical navigation. Public benchmarks have facilitated the development and evaluation of increasingly accurate registration methods. However, as registration errors approach the uncertainty of benchmark reference poses, measured performance may be limited by reference inaccuracies rather than algorithmic errors alone.

\textbf{Methods:} 
We propose Consensus Reference (ConRef), a multi-view framework for auditing reference poses in existing X-ray/computed tomography (CT) registration benchmarks. By jointly exploiting multiple X-ray observations, ConRef estimates an additional reference pose, quantifies its uncertainty, and identifies potential X-ray outliers. We validate the framework using independent anatomical evidence and a newly acquired X-ray/cone-beam
CT (CBCT) phantom dataset with bead-based reference poses.

\textbf{Results:} 
On our BeadPhant dataset, ConRef differed from independent bead-based references by 0.08--0.12\,mm, while discrepancies with original references reached 2.06\,mm on the Femur dataset. It also flagged 22 DeepFluoro X-rays as potential outliers. On Femur F5, reference choice changed the submillimeter success conclusions of the same registrations.

\textbf{Conclusion:} 
Our findings demonstrate that reference discrepancies can influence the evaluation of high-accuracy 2D/3D registration methods. By providing an additional, independently validated reference and identifying potential inconsistencies, ConRef enables systematic auditing of existing benchmarks and supports more reliable evaluation of registration error.
}


\maketitle

\section{Introduction}\label{sec:introduction}

Accurate medical image registration plays a fundamental role in image-guided surgery and surgical navigation by establishing spatial correspondences between images acquired at different stages, across imaging modalities, or in different coordinate systems. These correspondences enable preoperative anatomical information and surgical plans to be aligned with the intraoperative environment, supporting surgical navigation and image-guided interventions~\cite{li2025robotic}.

In orthopedic surgery, procedures such as spinal instrumentation, pelvic fixation, and complex fracture management require accurate localization of bony anatomy~\cite{li2026ultrasound}. Computed tomography (CT) and cone-beam CT (CBCT) provide detailed three-dimensional anatomical information, while intraoperative X-ray imaging provides two-dimensional projections for surgical guidance. Registering these complementary images enables preoperative 3D anatomy to be aligned with intraoperative observations. Developing and evaluating these 2D/3D registration methods requires datasets with reliable reference poses for quantitative evaluation.

Several public datasets, including DeepFluoro~\cite{grupp2020} and the Femur dataset~\cite{flepp2025}, have become widely used benchmarks for 2D/3D registration. Both datasets provide CT volumes, multiple corresponding C-arm X-rays, and reference poses for quantitative evaluation. They have supported the development and comparison of registration methods, including DiffPose~\cite{gopalakrishnan2024}, evaluated on DeepFluoro, and XVR~\cite{gopalakrishnan2026}, evaluated on both datasets. As registration methods become increasingly accurate, the reliability of benchmark reference poses becomes critical for interpreting their reported performance.

Reliable evaluation requires reference poses that are substantially more accurate than the registration methods being assessed~\cite{valstar2005,zhang2021}. Previous gold-standard datasets for 2D/3D registration have therefore explicitly quantified reference uncertainty~\cite{tomazevic2004,disidoro2021}. In practice, however, reference poses remain subject to  sources of uncertainty. DeepFluoro obtained its references through offline image registration followed by visual inspection~\cite{grupp2020}, whereas the Femur dataset derived them from a marker of metal spheres mounted on a K-wire~\cite{flepp2025}. When reported registration errors approach the millimeter or submillimeter scale, even small discrepancies in reference poses may affect the interpretation of registration error and threshold-based success criteria. Notably, the official XVR evaluation excludes four DeepFluoro X-rays due to questionable ground-truth pose parameters~\cite{xvrcode}. These observations highlight the importance of assessing reference reliability rather than treating the provided reference poses as unquestionable ground truth.

Although independent ground-truth measurements are often unavailable for existing datasets, their multiple X-ray views provide redundant geometric information that can be exploited to assess reference consistency. By jointly constraining the CT pose across multiple projections, it is possible to estimate an additional reference and examine whether X-rays agree with the resulting consensus. Motivated by this observation, we introduce Consensus Reference (ConRef), a multi-view framework for auditing reference poses in existing 2D/3D registration benchmarks. We further validate the framework using independent anatomical evidence and a newly acquired X-ray/CBCT phantom dataset with bead-based reference poses. Our contributions are:
\begin{itemize}
\item We propose ConRef, which provides an additional reference for assessing the reliability of existing 2D/3D registration benchmarks and quantifies its uncertainty.
\item We introduce an X-ray consistency assessment to identify potential X-ray outliers and demonstrate how reference discrepancies can affect the evaluation of registration error.
\item We provide the open-source framework and BeadPhant, an X-ray/CBCT phantom dataset with independently validated bead-based reference poses at a precision of the order of 0.1\,mm.\footnote{Code and dataset will be publicly released upon acceptance.}
\end{itemize}

\section{Method}\label{sec:method}

\begin{figure}[!t]
\centering
\includegraphics[width=\linewidth]{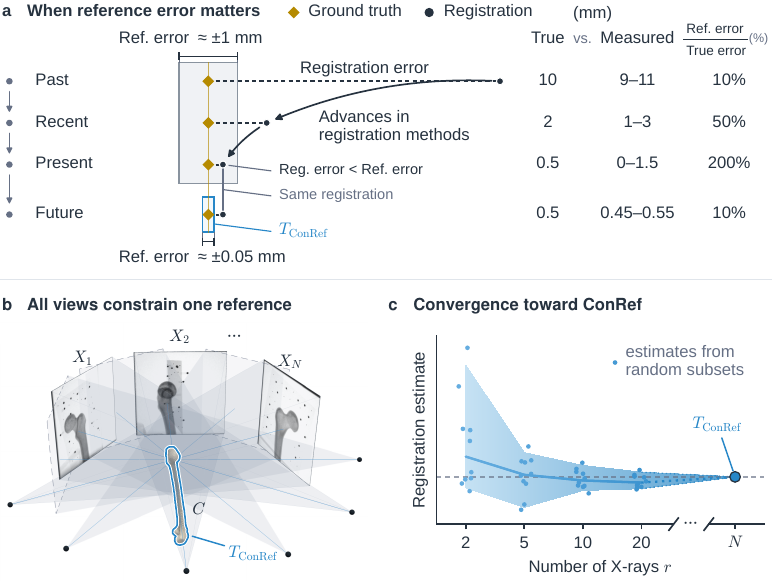}
\caption{
\textbf{Consensus reference (ConRef)}. (a) Reference uncertainty can distort measured registration error as methods improve (illustrative values). (b) Multiple X-rays with known relative projection geometries jointly constrain a single CT pose, $T_{\mathrm{ConRef}}$. (c) Estimates from random X-ray subsets approach the full-view ConRef as the number of views increases.
}
\label{fig:consensus}
\end{figure}

Our framework first estimates a consensus reference from multiple X-ray observations (Fig.~\ref{fig:consensus}). This reference is then used to assess dataset reliability by quantifying reference discrepancies and X-ray consistency, identifying potential anomalies, and evaluating registration results against an additional reference.

\subsection{Consensus Reference}

For each CT $C$, a multi-view dataset provides $N$ associated X-rays $\{X_i\}_{i=1}^{N}$ with known projection geometries, together denoted by $\{C, \{X_i\}_{i=1}^{N}\}$, and a reference pose $T_{\mathrm{OrigRef}}$ in a common coordinate frame. We refer to this pose as the original reference (OrigRef) rather than ground truth (GT), as its accuracy may be uncertain. Multiple X-ray views provide complementary geometric constraints for jointly estimating the CT pose. We define the consensus reference (ConRef) as the pose estimated from all available X-rays:

\begin{equation}
T_{\mathrm{ConRef}}
= \mathcal{R}\left(C,\{X_i\}_{i=1}^{N}\right),
\end{equation}
where $\mathcal{R}$ denotes a multi-view registration algorithm. To estimate ConRef, we optimize a single rigid transformation shared across all views while keeping their relative projection geometries fixed. Following our previous work~\cite{yue2026}, the registration jointly maximizes the average image similarity between the observed X-rays and the corresponding digitally reconstructed radiographs (DRRs):

\begin{equation}
T_{\mathrm{ConRef}} =
\arg\max_{T\in SE(3)}
\frac{1}{N}\sum_{i=1}^{N}
\mathcal{S}\left(X_i,\operatorname{DRR}_i(C,T)\right),
\end{equation}
where $\mathcal{S}$ denotes the image similarity metric, and $\operatorname{DRR}_i$ denotes DRR generation with DiffDRR~\cite{gopalakrishnan2022diffdrr} using the known projection geometry of the $i$-th X-ray.

We adopt the intensity- and gradient-based normalized cross-correlation (NCC) similarity from XVR~\cite{gopalakrishnan2026}, with equal weighting of both components. We replace the gradient-based optimizer with a Newton-based second-order optimizer that uses a locally fitted quadratic model and a coarse-to-fine image resolution strategy. Further implementation details are provided in the Supplementary Material.

ConRef thus provides an additional pose estimate supported by the available multi-view observations, although limited angular coverage and systematic modeling errors may still affect its accuracy.

\subsection{Dataset Assessment}

Based on ConRef, our framework can assess $\{C, \{X_i\}_{i=1}^{N}\}$.

\paragraph{CT reference-pose discrepancy.}

The discrepancy between the estimated ConRef and OrigRef can be quantified by $\Delta = d(T_{\mathrm{ConRef}}, T_{\mathrm{OrigRef}})$, where $d$ is the mean distance between the fiducials placed with the two poses. We define $d$ as the mean target reference distance (mTRD), following the mean target registration error (mTRE)~\cite{vandekraats2005}. The mTRE compares a registration result with GT, while the mTRD compares two references. A larger mTRD indicates a greater discrepancy between the two references.

In addition to the discrepancy $\Delta$, we assess the uncertainty of ConRef using randomly sampled X-ray subsets and a delete-$k$ jackknife~\cite{shao1989}. The resulting consensus uncertainty $U$ represents the estimated 95\% uncertainty radius due to finite X-ray observations, excluding shared systematic errors.

\paragraph{X-ray consistency.}
Although ConRef is estimated jointly from all X-rays, individual X-rays may differ in their consistency with ConRef. To assess this consistency, we perform single-view registration for each X-ray $X_i$, which optimizes the pose of $C$ on $X_i$ alone. The registration starts from $T_{\mathrm{ConRef}}$ as the initial pose and iteratively refines it until reaching the final pose $T_i$. If $X_i$ is consistent with ConRef, $T_i$ remains close to $T_{\mathrm{ConRef}}$. Otherwise, $T_i$ moves far from $T_{\mathrm{ConRef}}$. Since a single X-ray provides limited constraints on the position of $C$ along its rays, we measure the consistency using the mean reprojection distance (mRPD) $\rho_i$~\cite{vandekraats2005} between $T_i$ and $T_{\mathrm{ConRef}}$, which considers only differences perpendicular to the rays.

For each $\{C, \{X_i\}_{i=1}^{N}\}$, we report $\Delta$ and $U$, as well as the mean $\bar{\rho}$, standard deviation $\sigma_\rho$, and median of $\rho_i$, as the assessment results.

\subsection{Anomaly Detection}

Based on ConRef, our framework can identify potential anomalies in $\{C, \{X_i\}_{i=1}^{N}\}$.

\paragraph{CT reference-pose anomalies.}
For each CT, $\Delta$ measures the discrepancy between OrigRef and ConRef. We consider both the discrepancy $\Delta$ and consensus uncertainty $U$ to flag potentially anomalous discrepancies that warrant further inspection.

\paragraph{X-ray outliers.}
For each X-ray $X_i$, $\rho_i$ measures its consistency with ConRef. If $\rho_i$ stands out among the X-rays of its CT and also exceeds a threshold, we flag $X_i$ as a potential outlier that warrants further inspection. We identify outliers with the median-based rule of Leys et al.~\cite{leys2013}.

\subsection{Registration Evaluation}

Beyond dataset assessment, ConRef can also serve as a reference for evaluating registration results. For a registration result $T_{\mathrm{reg}}$, its mTRE is calculated as $d(T_{\mathrm{reg}}, T_{\mathrm{ConRef}})$.

By the triangle inequality, the mTREs of the same registration result against OrigRef and ConRef differ by at most $\Delta$. Such differences may affect whether a registration result reaches a submillimeter mTRE.

\section{Experiments and Results}\label{sec:experiments}
\subsection{Datasets}\label{sec:datasets}
We use three datasets: two publicly available datasets, Femur and DeepFluoro, and our newly released dataset, BeadPhant.

\begin{figure}[!t]
\centering
\includegraphics{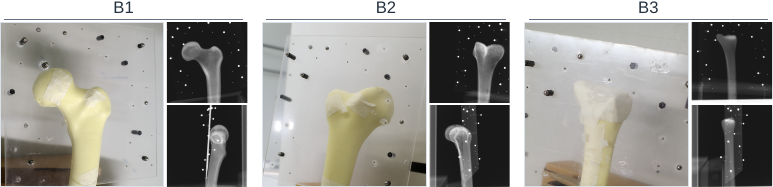}
\caption{The three CBCTs of BeadPhant. B1 and B2 show the same femur with the bead plate at its proximal and at its distal end, B3 a tibia. Each photo is shown with an X-ray taken from about the same direction (top) and one at a source angle of 90$^\circ$ (bottom).}\label{fig:beadref}
\end{figure}

\textbf{BeadPhant} is our phantom dataset for 2D/3D registration with high-precision OrigRef. It contains three CBCTs (B1--B3), two of the same synthetic femur and one of a synthetic tibia, fabricated as described in~\cite{watrinet2025} (Fig.~\ref{fig:beadref}). Each CBCT volume is associated with approximately 30 X-rays whose source angles are spread around a full circle. To make OrigRef as precise as possible, we deliberately use a phantom without soft tissue. The bone is fixed rigidly to an acrylic plate that carries steel beads, so the beads keep a stable position relative to the bone. The plate sits at the proximal end of the femur in B1, at its distal end in B2 and at the distal end of the tibia in B3. OrigRef is determined only by matching the beads in the CBCT to the beads in the X-rays. It involves no rendering or image similarity. An independent check with held-out beads shows a precision of the order of 0.1\,mm, measured as their median 3D error.

\textbf{Femur}~\cite{flepp2025} contains five CTs of cadaveric distal femurs (F1--F5) with 49 X-rays. Each CT volume is associated with approximately 10 X-rays acquired during a single C-arm sweep spanning 70--90$^\circ$. OrigRef was derived from a K-wire marker comprising 16 metal spheres mounted on a K-wire fixed to the femur. For F1--F3 and F5, an expert drew the femur contour on three of the X-rays. Two bone pins are also visible in the CT and in the X-rays.

\textbf{DeepFluoro}~\cite{grupp2020} contains six CTs of cadaveric pelvises (D1--D6) with 366 X-rays. Each CT volume is associated with 20 to more than 100 X-rays. OrigRef was obtained by offline single-view intensity-based registration of the pelvis followed by visual inspection to discard failed registrations.

\subsection{Consensus Reference Validation}
We validate ConRef against independent references of known precision. DeepFluoro remains valuable for the dataset assessment below, but its OrigRef rests on image similarity like ConRef and therefore cannot act as an independent reference. The OrigRef of BeadPhant is computed from the beads alone without image similarity, and its precision is of the order of 0.1\,mm. The OrigRef of Femur uses no image similarity either, but its known precision refers to the K-wire marker, while ConRef estimates the pose of the femur. To locate the femur itself, we fit an anatomical reference (AnatRef) to the expert contours and bone pins in the X-rays, again without image similarity (Supplementary Section~S2). Only on F5 is AnatRef precise to about 0.2\,mm in every direction, so we use it only for F5.

Table~\ref{tab:validation} compares ConRef with the four independent references. The observed discrepancies are comparable to the estimated precision of the independent references, supporting agreement between ConRef and these references at the submillimeter scale.

\begin{table}[!t]
\caption{mTRD between ConRef and each independent reference, with the precision of that reference. Precision is the median 3D error of held-out beads for BeadPhant and the median mTRD of poses drawn from the uncertainty of AnatRef for F5.}\label{tab:validation}
\centering
\begin{tabular}{@{}llcc@{}}
\toprule
CT & Reference & mTRD to ConRef (mm) & Precision of reference (mm) \\
\midrule
B1 & Bead-based OrigRef & 0.12 & 0.08 \\
B2 & Bead-based OrigRef & 0.08 & 0.05 \\
B3 & Bead-based OrigRef & 0.11 & 0.17 \\
F5 & AnatRef & 0.30 & 0.27 \\
\botrule
\end{tabular}

\end{table}

\subsection{Dataset Assessment and Anomaly Detection}

To estimate consensus uncertainty $U$, we randomly select $r=\lfloor N/2\rfloor$ X-rays per CT and repeat the multi-view registration ten times. We then estimate the full-view pose covariance using a delete-$k$ jackknife correction and derive $U$ as the 95th-percentile mTRD under a Gaussian approximation of the pose distribution.

For anomaly screening, we flag the discrepancy when $\Delta+U$ reaches or exceeds 0.5\,mm. An X-ray is flagged when its mRPD exceeds both 1\,mm and the median plus three scaled median absolute deviations (MADs) of the corresponding CT.

\begin{figure}[!t]
\centering
\includegraphics{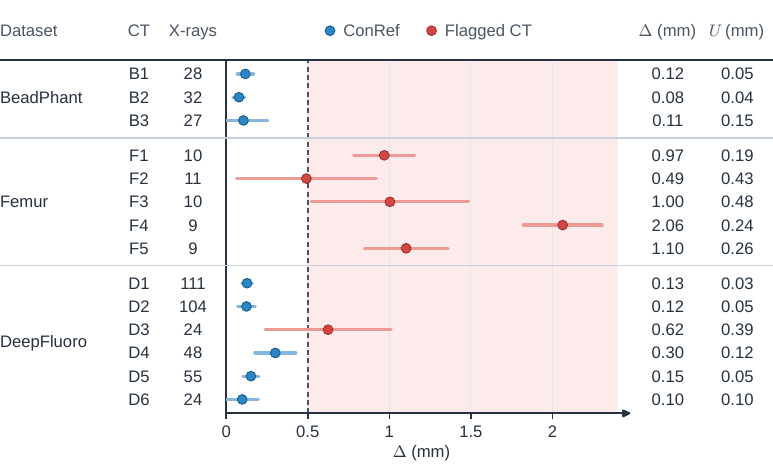}
\caption{CT reference-pose discrepancy and anomalies. ConRef of every CT at its distance $\Delta$ to OrigRef, which lies at zero. Horizontal bars span $\Delta \pm U$, where $U$ is the uncertainty of ConRef. The discrepancy is flagged when its bar reaches the dashed line at 0.5\,mm.}\label{fig:reference-distance}
\end{figure}

\paragraph{CT reference-pose discrepancy and anomalies.}
Fig.~\ref{fig:reference-distance} shows discrepancy $\Delta$ and consensus uncertainty $U$ for each $\{C, \{X_i\}_{i=1}^{N}\}$. 

OrigRef and ConRef differed little on BeadPhant and most DeepFluoro CTs, whereas larger discrepancies were observed on Femur. 

The discrepancies were flagged as anomalous for all Femur CTs and only D3 in DeepFluoro.

\begin{figure}[!t]
\centering
\includegraphics{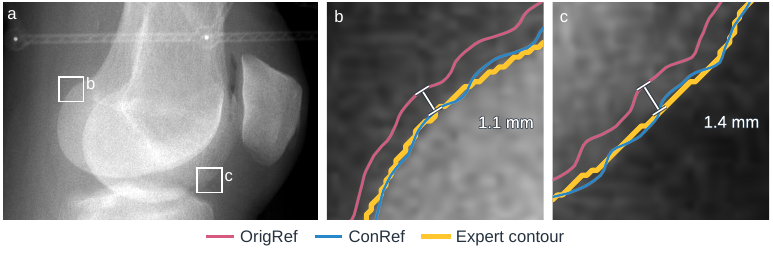}
\caption{Femur outlines on an X-ray of F5. Boxes in (a) mark the regions enlarged in (b) and (c). The femur in the CT is projected onto the X-ray with OrigRef and with ConRef. The ConRef outline lies closer to the expert contour. White bars mark the distance between the OrigRef and ConRef outlines.}\label{fig:femur-contours}
\end{figure}

\begin{figure}[!t]
\centering
\includegraphics[width=\linewidth]{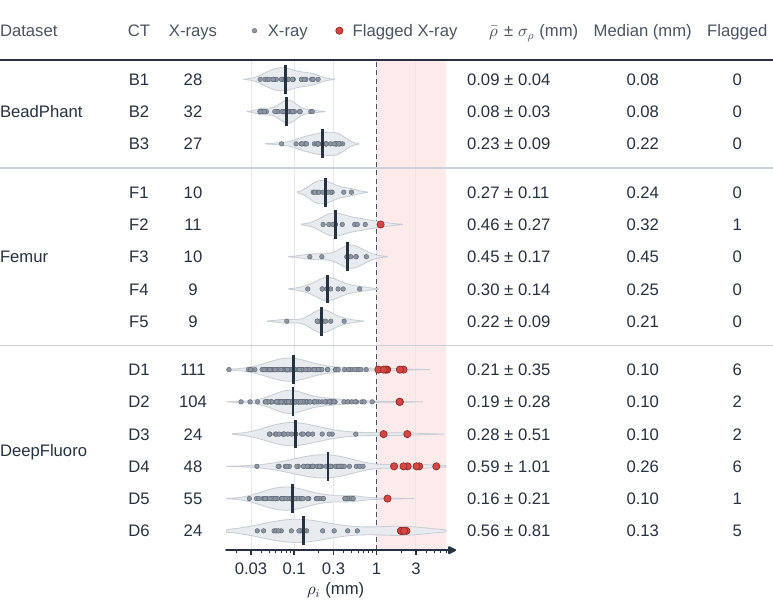}
\caption{X-ray consistency and outliers. $\rho_i$ of every X-ray, one row per CT. The grey shape of each row widens where X-rays are dense. Vertical bars mark the median of each CT. The dashed line marks the 1\,mm threshold, which a flagged X-ray must exceed in addition to the median-based rule, for all three datasets. The horizontal axis is logarithmic. The columns on the right list $\bar{\rho} \pm \sigma_\rho$, the median of $\rho_i$ and the number of flagged X-rays of each CT. Supplementary Table~S2 identifies all red points.}\label{fig:consistency}
\end{figure}

\paragraph{X-ray consistency and outliers.}
Fig.~\ref{fig:consistency} shows the distribution of $\rho_i$, the mean $\bar{\rho}$, standard deviation $\sigma_\rho$, and median of $\rho_i$ for each $\{C, \{X_i\}_{i=1}^{N}\}$. 

Median $\rho_i$ values were generally low across all three datasets, but DeepFluoro exhibited some substantially larger $\rho_i$ values. 

No outliers were flagged on BeadPhant, compared with one on Femur and 22 on DeepFluoro (listed in Supplementary Table~S2). The largest $\rho_i$ was 5.30\,mm (X-ray 002 of D4).

\paragraph{Interpretation of the flags.}
For the Femur CTs, OrigRef follows the K-wire marker and ConRef follows the femur itself. The Femur discrepancies in Fig.~\ref{fig:reference-distance} therefore may partly be due to the relative pose of the femur and marker differing slightly between CT and X-ray imaging. Fig.~\ref{fig:femur-contours} shows this on F5. The femur segmented in the CT is projected onto one X-ray, once with OrigRef and once with ConRef. The expert contour was drawn on the X-ray itself and depends on neither reference, and the ConRef outline lies closer to it.

In DeepFluoro, the discrepancy of D3 lies mainly along the rays. Because the X-rays of D3 span the narrowest range of viewing directions, X-rays constrain the CT pose least in this direction.

Among the flagged DeepFluoro X-rays in Fig.~\ref{fig:consistency} are three of the four that the XVR evaluation excludes~\cite{gopalakrishnan2026,xvrcode}.

\subsection{Registration Evaluation}

This experiment has two aims: to demonstrate the use of ConRef as an additional reference for evaluating registration error, and to address potential circularity when evaluating methods that share the objective used to construct ConRef. For the first aim, we evaluate each registration of our previous intensity-based method~\cite{yue2026}, which shares the objective of ConRef, against ConRef, OrigRef and AnatRef. To address the second aim, we also evaluate our reimplementation of the method of Flepp et al.~\cite{flepp2025}, which matches automatically extracted bone contours using a different objective from the intensity-based registration.

We demonstrate this on F5, where AnatRef is available. On BeadPhant, OrigRef and ConRef differ by only about 0.1\,mm, so a registration has nearly the same mTRE against either of them. We performed two-view registration with both methods on all 36 pairs formed from the nine X-rays of F5. 

\begin{figure}[!htb]
\centering
\includegraphics{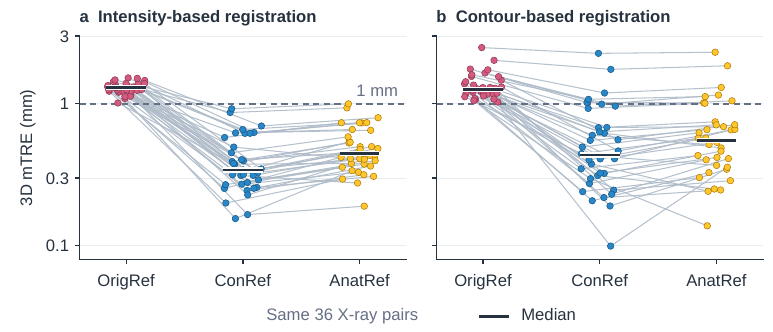}
\caption{Registration errors for the same 36 X-ray pairs of F5 using our previous intensity-based method~\cite{yue2026} (a) and our contour-based reimplementation of Flepp et al.~\cite{flepp2025} (b), which does not use intensity-based image similarity. Each thin line connects the mTREs of one registration against OrigRef, ConRef and AnatRef. Black horizontal bars mark medians. The dashed line marks 1\,mm. The vertical axis is logarithmic.}\label{fig:femur-registrations}
\end{figure}

The results are shown in Fig.~\ref{fig:femur-registrations} and address both aims:
First, ConRef can be used to evaluate registration error. Its assessments largely agree with those of the independent AnatRef. Second, the agreement between ConRef- and AnatRef-based evaluations for the contour-based method further reduces concerns about objective-dependent evaluation bias. More importantly, OrigRef would lead to a different conclusion: against OrigRef, no registration falls below 1\,mm, as if neither method could reach a submillimeter mTRE. With ConRef and AnatRef, both methods achieve submillimeter 3D registration.
\section{Discussion}\label{sec:discussion}

The reliability of benchmark reference poses has also been questioned in computer vision. Brachmann et al.~\cite{brachmann2021} demonstrated that different pseudo-ground-truth generation methods can alter the evaluation and ranking of visual camera re-localization methods. Similar concerns arise in 2D/3D medical image registration, where reference discrepancies may affect reported registration error. ConRef addresses this issue by providing an additional reference for benchmark assessment and identifying potential X-ray outliers. Rather than automatically correcting existing annotations, the framework highlights cases that warrant further inspection.

ConRef is not restricted to a particular registration algorithm $\mathcal R$. In this study, we implemented the framework using intensity-based multi-view registration and validated the resulting references against independently constructed bead-based and anatomical references. Our BeadPhant dataset provides a controlled validation setting, with CBCT and X-rays acquired using the same imaging system and reference poses established independently through rigidly attached beads. The agreement with these references supports the reliability of ConRef in the evaluated configurations. Nevertheless, different registration algorithms may produce different consensus estimates, and their potential systematic biases cannot be fully excluded.

The reliability of ConRef also depends on the available X-ray observations. Limited angular coverage provides weaker constraints on certain pose components, particularly translation along the rays. Moreover, consensus uncertainty measures variability due to finite-view sampling rather than absolute ground-truth accuracy, and shared errors in projection geometry or registration models may remain undetected. These limitations should be considered when applying ConRef to other datasets or using it as an additional evaluation reference.

\section{Conclusion}\label{sec:conclusion}

We introduced Consensus Reference (ConRef), a multi-view framework for auditing reference reliability in 2D/3D registration benchmarks. By leveraging multiple X-ray observations, ConRef provides an additional reference for assessing pose discrepancies, estimating consensus uncertainty, and identifying potential X-ray outliers. Independent validation using bead-based and anatomical references supports its reliability in the evaluated configurations. Our results demonstrate that benchmark reference discrepancies can substantially affect reported registration error and even change submillimeter success conclusions. ConRef therefore provides a practical approach to assessing existing benchmarks and improving the interpretation of high-accuracy registration results. We also provide the framework and BeadPhant dataset to support future research.

\backmatter

\bibliography{sn-bibliography}

\end{document}


\title{Supplementary Material\break ConRef: Auditing Reference Reliability in 2D/3D Registration Benchmarks}
\maketitle

\section{Consensus Reference (ConRef)}\label{sec:s-conref}
The relative projection geometry of the X-rays of a CT is taken from OrigRef. Half of the similarity $\mathcal{S}$ of XVR~\cite{gopalakrishnan2026} is the normalized cross-correlation of intensities, over the whole image and over $9\times9$ patches with equal weights, and half is the normalized cross-correlation of gradients over $11\times11$ patches. Only the bone of interest is rendered, the femur on Femur, the femur or the tibia on BeadPhant and the pelvic labels that XVR uses on DeepFluoro. Each dataset uses the preprocessing of XVR, but the DeepFluoro X-rays are not cropped. BeadPhant uses the preprocessing of Femur. We set DiffDRR to sample each voxel at its center. The OrigRef of each DeepFluoro CT was adapted to this convention by a fixed translation of $(0.5, 0.5, -0.5)$\,mm. A Newton method fits a quadratic model of the similarity to renderings at 85 transformations around the current estimate and steps toward its maximum. A step is kept only if the similarity increases. Otherwise, shorter steps in the same direction are tried. The allowed step length grows when a kept step uses all of it and shrinks when no step is kept or only a much shorter one. The method stops when a kept step moves the fiducials by less than 0.01\,mm on average or when the allowed step length falls below 0.005\,mm. The registration runs at a quarter of the native image resolution and then, on Femur and BeadPhant, at the native resolution. On DeepFluoro, a further stage at the native resolution moved the result by less than 0.05\,mm on the two CTs tested. All registrations for ConRef, including those on the random subsets below, start from OrigRef. Starting instead from OrigRef moved by 1\,mm and 1$^\circ$ or by 2\,mm and 2$^\circ$ changed ConRef by less than 0.05\,mm. The single-view registration for the mRPD is the same method applied to one X-ray and starts from ConRef.

For $U$, the covariance of the ten subset estimates around ConRef in the six rigid-motion parameters is scaled by the delete-$k$ jackknife factor $r/k$, where $k = N - r$ is the number of X-rays left out~\cite{shao1989}. $U$ is the 95th percentile of the mTRDs of rigid motions drawn from a Gaussian distribution with this covariance.

\section{Anatomical Reference}\label{sec:s-anatomical}
The femur transformation estimates are built only from geometry and expert annotation, without image similarity, DRRs or registration results. The cameras are re-estimated from the metal spheres of the marker, with one focal length per CT. The unknown is one rigid transform of the femur relative to the marker per CT, so OrigRef is the special case of no motion. Two kinds of observations constrain it:
\begin{itemize}
\item \textbf{Contours.} The expert femur contours that the dataset provides on three X-rays per CT~\cite{flepp2025} are matched to the projected outline of the femur surface.
\item \textbf{Bone pins.} The shadows of the two pins inside the bone are traced along each pin in every X-ray.
\end{itemize}
Contours and pins are first fitted separately and tested for agreement. A pin is kept only if it agrees with the contours. The uncertainty of the resulting transformation combines a bootstrap over X-rays and contour segments, the uncertainty of the re-estimated cameras and the effect of modeling choices, and it was calibrated on synthetic data with known motion. Table~\ref{tab:s-anatomical} lists the evidence and uncertainty of the estimate for each CT. In F3 and F5, one of the two pins agrees with the contours and joins them under a rule fixed in advance. In F1 and F2, the two pins together disagree with the contours and the rule fixed in advance does not resolve this. The decision to use the contours alone was made after a first comparison with ConRef. F4 has no estimate, because contour annotations are provided only for the other four CTs and its two pins disagree. With contours on three X-rays, one direction is determined more loosely than the others, most noticeably in F1 and F3, with standard deviations of 0.65 and 0.84\,mm. F5 is the only CT that meets all three conditions: two kinds of evidence, a rule fixed in advance and a standard deviation of at most 0.23\,mm in every direction. Its estimate therefore serves as the anatomical reference (AnatRef) in the main text.

\begin{table}[h]
\caption{Femur transformation estimates from contours and pins. Evidence, whether the rule for combining it was fixed in advance, and standard deviation (mm) of the femur displacement along the least and the second least certain direction. The estimate of F5 serves as AnatRef.}\label{tab:s-anatomical}
\centering
\begin{tabular}{@{}llccc@{}}
\toprule
CT & Evidence & Fixed in advance & Least certain & Second least certain \\
\midrule
F1 & Contours & No & 0.65 & 0.26 \\
F2 & Contours & No & 0.45 & 0.07 \\
F3 & Contours and one pin & Yes & 0.84 & 0.27 \\
F4 & None & -- & -- & -- \\
F5 & Contours and one pin & Yes & 0.23 & 0.09 \\
\botrule
\end{tabular}
\end{table}

Because AnatRef uses re-estimated cameras, it is compared with the other references through the motion of the femur relative to the marker, which does not depend on the cameras. Its calibrated uncertainty is a Gaussian distribution over the six rigid-motion parameters of the femur. The precision of AnatRef in the main text is the median mTRD between AnatRef and 2000 rigid transformations drawn from this distribution.

\section{Flagged X-rays}\label{sec:s-flagged}
Table~\ref{tab:s-flagged-df} identifies all 23 red points in the X-ray consistency figure of the main text: one flagged Femur X-ray and 22 flagged DeepFluoro X-rays. No BeadPhant X-ray was flagged. Each point is identified by its dataset, CT and original X-ray ID, retaining leading zeros.

Three of the flagged DeepFluoro X-rays are among the four X-rays that the XVR evaluation excludes~\cite{gopalakrishnan2026,xvrcode}. The evaluation script of XVR links to a note by its first author in the dataset repository that the projected segmentations of these X-rays do not line up with the images. The fourth, D1 X-ray 050, has an mRPD of 0.50\,mm. The flagged Femur X-ray, F2 X-ray 04, has an mRPD of 1.12\,mm, against 0.22--0.73\,mm for the other ten X-rays of this CT.

\begin{table}[h]
\caption{All flagged X-rays shown as red points in the X-ray consistency figure of the main text, with their mRPD to ConRef. $^{*}$\,X-rays that the XVR evaluation excludes.}\label{tab:s-flagged-df}
\centering
\begin{tabular}{@{}lccc@{}}
\toprule
Dataset & CT & X-ray & mRPD (mm) \\
\midrule
Femur & F2 & 04 & 1.12 \\
\midrule
DeepFluoro & D1 & 003$^{*}$ & 2.12 \\
 & D1 & 084 & 1.35 \\
 & D1 & 085 & 1.29 \\
 & D1 & 088 & 1.92 \\
 & D1 & 096 & 1.05 \\
 & D1 & 097 & 1.22 \\
 & D2 & 016 & 1.92 \\
 & D2 & 034 & 1.90 \\
 & D3 & 009 & 2.36 \\
 & D3 & 016 & 1.21 \\
 & D4 & 000 & 3.28 \\
 & D4 & 002$^{*}$ & 5.30 \\
 & D4 & 004$^{*}$ & 2.38 \\
 & D4 & 022 & 1.63 \\
 & D4 & 029 & 2.12 \\
 & D4 & 038 & 3.05 \\
 & D5 & 054 & 1.36 \\
 & D6 & 000 & 2.30 \\
 & D6 & 019 & 1.98 \\
 & D6 & 021 & 2.04 \\
 & D6 & 022 & 1.98 \\
 & D6 & 023 & 2.15 \\
\botrule
\end{tabular}

\end{table}

\bibliography{sn-bibliography}